\documentclass[10pt,journal,compsoc]{IEEEtran}
\ifCLASSOPTIONcompsoc
  \usepackage[nocompress]{cite}
\else
  \usepackage{cite}
\fi
\ifCLASSINFOpdf
\else
\fi
\usepackage{epsfig}
\usepackage{graphicx}
\usepackage{amsmath}
\usepackage{amssymb}
\usepackage{multirow}
\usepackage{upgreek}
\usepackage[table,xcdraw,dvipsnames]{xcolor}
\usepackage[pagebackref=true,breaklinks=true,colorlinks,bookmarks=false]{hyperref}

\begin{document}
\title{TypeNet: Deep Learning Keystroke Biometrics}
% lists the Biometrics Council journals use for "first footnote" author

\author{Alejandro Acien, Aythami Morales, John V. Monaco, Ruben Vera-Rodriguez, Julian Fierrez,~\IEEEmembership{Member,~IEEE}% <-this % stops a space

\IEEEcompsocitemizethanks{\IEEEcompsocthanksitem A. Acien, A. Morales, R. Vera-Rodriguez, and J. Fierrez are with the School of Engineering, Universidad Autonoma de Madrid, 28049 Madrid, Spain (e-mail: alejandro.acien@uam.es; aythami.morales@uam.es; ruben.vera@uam.es; julian.fierrez@uam.es).
\IEEEcompsocthanksitem J. V. Monaco is with the Naval Postgraduate School, Monterey CA, USA (e-mail: vinnie.monaco@nps.edu).}}% <-this % stops a space



\markboth{Journal of \LaTeX\ Class Files,~Vol.~14, No.~8, February~2021}%
{Shell \MakeLowercase{\textit{et al.}}: Bare Demo of IEEEtran.cls for Biometrics Council Journals}

% Biometrics Council journal papers as those publications place the marks
% or like this to get the Biometrics Council new two part style.
%\hspace{\columnsep}\makebox[\columnwidth]{Published by the IEEE Biometrics Council\hfill}}
% column for its text to clear the IEEEpubid mark (Biometrics Council jorunal


% for Biometrics Council papers, we must declare the abstract and index terms
\IEEEtitleabstractindextext{%
\begin{abstract}
   We study the performance of Long Short-Term Memory networks for keystroke biometric authentication at large scale in free-text scenarios. For this we \textcolor{black}{explore the performance of Long Short-Term Memory (LSTMs) networks} trained with a moderate number of keystrokes per identity \textcolor{black}{and evaluated under different scenarios including: i) three learning approaches depending on the loss function (softmax, contrastive, and triplet loss); ii) different number of training samples and lengths of keystroke sequences; iii) four databases based on two device types (physical vs touchscreen keyboard); and iv) comparison with existing approaches based on both traditional statistical methods and deep learning architectures. Our approach called} TypeNet achieves state-of-the-art keystroke biometric authentication performance with an Equal Error Rate of 2.2\% and 9.2\% for physical and touchscreen keyboards, respectively, significantly outperforming previous approaches. Our experiments demonstrate a moderate increase in error with up to 100,000 subjects, demonstrating the potential of TypeNet to operate at an Internet scale. To the best of our knowledge, \textcolor{black}{the databases used in this work} are the largest existing free-text keystroke databases available for research with more than 136 million keystrokes from 168,000 subjects in physical keyboards, and 60,000 subjects with more than 63 million keystrokes acquired on mobile touchscreens.
\end{abstract}

\begin{IEEEkeywords}
Biometrics, keystroke dynamics, large scale, deep learning, TypeNet, keystroke authentication.
\end{IEEEkeywords}}

\maketitle

\IEEEdisplaynontitleabstractindextext

% \fi
\IEEEpeerreviewmaketitle

\IEEEraisesectionheading{\section{Introduction}\label{sec:introduction}}
Keystroke dynamics is a behavioral biometric trait aimed at recognizing individuals based on their typing habits. The velocity of pressing and releasing different keys \cite{Banerjee}, the hand postures during typing \cite{Buschek}, and the pressure exerted when pressing a key \cite{10.1007/978-3-030-31321-0_2} are some of the features taken into account by keystroke biometric algorithms aimed to discriminate among subjects. Although keystroke biometrics suffer high intra-class variability for person recognition, especially in free-text scenarios (i.e. the input text typed is not fixed between enrollment and testing), the ubiquity of keyboards as a method of text entry makes keystroke dynamics a near universal modality to authenticate subjects on the Internet.

Text entry is prevalent in day-to-day applications: unlocking a smartphone, accessing a bank account, chatting with acquaintances, email composition, posting content on a social network, and e-learning \cite{2020_AAAI_edBB_JH}. As a means of subject authentication, keystroke dynamics is economical because it can be deployed on commodity hardware and remains transparent to the user. These properties have prompted several companies to capture and analyze keystrokes. The global keystroke biometrics market is projected to grow from \$$129.8$ million dollars (2017 estimate) to \$$754.9$ million by 2025, a rate of up to $25\%$ per year\footnote{\href{https://www.prnewswire.com/news-releases/keystroke-dynamics-market-to-reach-754-9-mn-globally-by-2025-at-24-7-cagr-says-amr-300790697.html}{https://www.prnewswire.com/news-releases/keystroke}}. As an example, Google has recently committed \$7 million dollars to fund TypingDNA\footnote{\href{https://siliconcanals.com/news/google-leads-funding-typingdna-authenticate-typing-behaviour/}{https://siliconcanals.com/news/}}, a startup company which authenticates people based on their typing behavior.

At the same time, the security challenges that keystroke biometrics promises to solve are constantly evolving and getting more sophisticated every year: identity fraud, account takeover, sending unauthorized emails, and credit card fraud are some examples\footnote{\href{https://150sec.com/fraudulent-fingertips-how-typing-biometrics-are-changing-cybersecurity/13466/}{https://150sec.com/fraudulent-fingertips}}. These challenges are magnified when dealing with applications that have hundreds of thousands to millions of users. In this context, keystroke biometric algorithms capable of authenticating individuals while interacting with online applications are more necessary than ever. As an example of this, Wikipedia struggles to solve the problem of \textit{`edit wars'} that happens when different groups of editors \textcolor{Black}{representing opposite opinions undo their changes reciprocally in an attempt to impose their version}. According to \cite{Yasseri}, up to $12$\% of the discussions in Wikipedia are devoted to revert changes and vandalism, suggesting that the Wikipedia criteria to identify and resolve controversial articles is highly contentious. Keystroke biometrics algorithms could be used to identify these malicious editors among the thousands of editors who write articles in Wikipedia every day. Other applications of keystroke biometric technologies are found in e-learning platforms; student identity fraud and cheating are some challenges that virtual education technologies need to addresss to become a viable alternative to face-to-face education \cite{2020_AAAI_edBB_JH}.  

The literature on keystroke biometrics is extensive, but to the best of our knowledge, previous systems have only been evaluated with up to several hundred subjects and cannot deal with the recent challenges that massive usage applications are facing. The aim of this paper is to explore the feasibility and limits of deep learning architectures for scaling up free-text keystroke biometrics to hundreds of thousands of users. The main contributions of this work are threefold: 

\begin{enumerate}
\setlength\itemsep{0em}
\item We explore novel free-text keystroke biometrics approaches based on Deep Recurrent Neural Networks, suitable for authentication and identification at large scale. We conduct an exhaustive experimentation and evaluate how performance is affected by the following factors: the length of the keystroke sequences, the number of gallery samples, and the device (touchscreen vs physical keyboard). We present TypeNet, a Recurrent Neural Network trained with keystroke sequences from more than $100$,$000$ subjects. We analyze the performance of three different \textcolor{black}{learning strategies based on traditional classification frameworks (softmax) and Distance-metric approaches (contrastive and triplet loss).} 

\item The results reported by TypeNet represent the state of the art in keystroke authentication based on free-text reducing the error obtained by previous works in more than 50\%. Processed data has been made available so the results can be reproduced\footnote{Data available at: \url{https://github.com/BiDAlab/TypeNet}}. We evaluate TypeNet in terms of Equal Error Rate (EER) as the number of test subjects is scaled from $100$ up to $100$,$000$ (independent from the training data) for the desktop scenario (physical keyboards) and up to $30$,$000$  for the mobile scenario (touchscreen keyboard). TypeNet learns a feature representation of a keystroke sequence without the need for retraining if new subjects are added to the database, as commonly happens in many biometric systems \cite{Fierrez-Aguilar2005_AdaptedMultimodal}. Therefore, TypeNet is easily scalable.

\item We carry out a comparison with previous state-of-the-art approaches for free-text keystroke biometric authentication. \textcolor{black}{Our experiments include four state-of-the-art approaches and four public databases.} The performance achieved by the proposed method outperforms previous approaches in the scenarios evaluated in this work. The results suggest that authentication error rates achieved by TypeNet remain low as thousands of new users are enrolled. 
\end{enumerate}

A preliminary version of this article was presented in \cite{TypeNet}. This article significantly improves \cite{TypeNet} in the following aspects:
\begin{enumerate}
\setlength\itemsep{0em}
\item We add a new version of TypeNet trained and tested with keystroke sequences acquired in mobile devices and results in the mobile scenario. Additionally, we provide cross-sensor interoperability results \cite{alonso10qualityBasedSensorInteroperability, 2017_PLOSONE_eBioSign_Tolosana} between desktop and mobile datasets.
\item We include two new \textcolor{black}{learning strategies based on} softmax and triplet loss, that serve to improve the performances in all scenarios. Our experiments demonstrate that triplet loss can be used to multiply by two the accuracy of free-text keystroke authentication approaches. 
\item We evaluate TypeNet in terms of Rank-{n} identification rates using a background set of $1$,$000$ subjects (independent from the training data).
\item \textcolor{black}{We also test our TypeNet models with two \textcolor{black}{additional} state-of-the-art free-text keystroke datasets, demonstrating the potential of our method to generalize well with other databases.}
\item We add experiments about the dependencies between input text and TypeNet performance, a common issue in free-text keystroke biometrics.
\end{enumerate}

In summary, we present the first evidence in the literature of competitive performance of free-text keystroke biometric authentication at large scale (up to $100$,$000$ test subjects). The results reported in this work demonstrate the potential of this behavioral biometric for widespread deployment.

The paper is organized as follows: Section \ref{related_works} summarizes related works in free-text keystroke dynamics. Section \ref{aalto} describes the datasets used for training and testing TypeNet models. Section \ref{system_description} describes the processing steps and learning methods in TypeNet. Section \ref{experimental_protocol} details the experimental protocol. Section \ref{experiments_results} reports the experiments and discusses the results obtained. Section \ref{conclusions} summarizes the conclusions and future work.

%-------------------------------------------------------------------------
\section{Background and Related Work}
\label{related_works}

\begin{table*}
\begin{center}
\begin{tabular}{|c|c|c|c|c|c|c|}
\hline
Study & Scenario & \#Subjects  & \#Seq.  & Sequence Size & \#Keys & Best Performance\\
\hline\hline
Monrose and Rubin (1997) \cite{Monrose} & Desktop & $31$	& N/A     & N/A	  & N/A & ACC = $23$\%\\
Gunetti and Picardi (2005) \cite{Gunetti} & Desktop & $205$	& $1$$\sim$$15$ &	$700$$\sim$$900$ keys &	$688$K & EER = $7.33$\%\\
Kim and Kang  (2009) \cite{kim2020freely} & Mobile & $50$	& $20$   & $\sim$$200$ keys    & $200$K & EER = $0.05$\%\\
Gascon \textit{et al.} (2014) \cite{gascon2014continuous} & Mobile & $315$	& $1$$\sim$$10$   & $\sim$$160$ keys    & $67$K & EER = $10.0$\%\\

Ceker and Upadhyaya (2016) \cite{Ceker}   & Desktop & $34$	& $2$   &	$\sim$$7$K keys       & $442$K & EER = $2.94$\% \\
Murphy \textit{et al.} (2017) \cite{Murphy}  & Desktop & $ 103$	& N/A     &	$1$,$000$ keys	            & $12.9$M & EER = $10.36$\%\\
Monaco and Tappert (2018) \cite{Monaco}  & Both & $55$	& $6$         &	$500$ keys	            & $165$K & EER = $0.6$\%\\
\textcolor{black}{Lu \textit{et al.} (2019) \cite{Lu}} & Desktop & $75$	& $3$   & $\sim$$5$,$700$ keys    & $1$,$2$M & EER = $3.04$\%\\

Deb \textit{et al.} (2019) \cite{Deb}     & Mobile & $37$	& $180$K      &	3 seconds	            & $6.7$M & $81.61$\% TAR at $0.1$\% FAR\\

\textbf{Ours (2020)}           & Both & \boldmath$228$K	& \boldmath$15$        &	 \boldmath $\sim$$70$ \textbf{keys}	        & \boldmath$199$\textbf{M} & \textbf{EER =} \boldmath$2.2\%$\\
\hline
\end{tabular}
\end{center}
\caption{Comparison among different free-text keystroke datasets employed in relevant related works. N/A = Not Available. ACC = Accuracy, EER = Equal Error Rate, TAR = True Acceptance Rate, FAR = False  Acceptance Rate.}
\label{table:works}
\end{table*}

The measurement of keystroke dynamics depends on the acquisition of key press and release events. This can occur on almost any commodity device that supports text entry, including desktop and laptop computers, mobile and touchscreen devices that implement soft (virtual) keyboards, and PIN entry devices such as those used to process credit card transactions. Generally, each keystroke (the action of pressing and releasing a single key) results in a keydown event followed by keyup event, and the sequence of these timings is used to characterize an individual's keystroke dynamics. Within a web browser, the acquisition of keydown and keyup event timings requires no special permissions, enabling the deployment of keystroke biometric systems across the Internet in a transparent manner.

Keystroke biometric systems are commonly placed into two categories: \textit{fixed-text}, where the keystroke sequence typed by the subject is prefixed, such as a username or password, and \textit{free-text}, where the keystroke sequence is arbitrary, such as writing an email or transcribing a sentence with typing errors. Notably, free-text input results in different keystroke sequences between the gallery and test samples as opposed to fixed-text input. Biometric authentication algorithms based on keystroke dynamics for desktop and laptop keyboards have been predominantly studied in fixed-text scenarios where accuracies higher than $95\%$ are common \cite{2016_IEEEAccess_KBOC_Aythami}. Approaches based on sample alignment (e.g. Dynamic Time Warping) \cite{2016_IEEEAccess_KBOC_Aythami}, Manhattan distances \cite{Vinnie1}, digraphs \cite{Bergadano}, and statistical models (e.g. Hidden Markov Models) \cite{Ali} have shown to achieve the best results in fixed-text.

Nevertheless, the performances of free-text algorithms are generally far from those reached in the fixed-text scenario, where the complexity and variability of the text entry contribute to intra-subject variations in behavior, challenging the ability to recognize subjects \cite{Sim}. Murphy \textit{et al.} \cite{Murphy} collected a very large free-text keystroke dataset ($\sim$$12.9$M keystrokes) called Clarkson II and applied the Gunetti and Picardi algorithm \cite{Gunetti} achieving $10.36\%$ classification error using sequences of $1$,$000$ keystrokes and $10$ genuine sequences to authenticate subjects. \textcolor{black}{The effect of the data size on the performance of free-text keystroke algorithms has been studied by Huang \textit{et al.} \cite{Huang_2015}. Their results suggested that a sample size of $10$,$000$ keystrokes for the reference profile and $1$,$000$ keystrokes for the test sample are needed to achieve good authentication performance for those algorithms based on n-graph features. The main drawback when using large keystroke sequences was that the subject needed on average six minutes of typing to generate a valid sample. Finally, in \cite{ayotte2020fast} the authors implemented a new metric based on Random Forest classifier to select the best features for keystroke recognition when using digraph algorithms. Their results on the Clarkson II \cite{Murphy} dataset achieved a $7.8\%$ EER with $200$ digraphs, demonstrating the potential of such algorithms with an appropriate selection of the keystroke features.}

More recently than the pioneering works of Monrose and Gunetti, some algorithms based on statistical models have shown to work very well with free-text, like the POHMM (Partially Observable Hidden Markov Model) \cite{Monaco}. This algorithm is an extension of the traditional Hidden Markov Model (HMM), but with the difference that each hidden state is conditioned on an independent Markov chain. This algorithm is motivated by the idea that keystroke timings depend both on past events and the particular key that was pressed. Performance achieved using this approach in free-text is close to fixed-text, but it again requires several hundred keystrokes and has only been evaluated with a database containing less than $100$ subjects. 

Because mobile devices are not stationary, mobile keystroke biometrics depend more heavily on environmental conditions, such as the user's location or posture, than physical keyboards which typically remain stationary \cite{palin}. This challenge of mobile keystroke biometrics was examined by Crawford and Ahmadzadeh in \cite{crawford2017authentication}. They found that authenticating a user in different positions (sitting, standing, or walking)  performed only slightly better than guessing, but detecting the user's position before authentication can significantly improve performance.

Like desktop keystroke biometrics, many mobile keystroke biometric studies have focused on fixed-text sequences \cite{teh2016survey}. Some recent works have considered free-text sequences on mobile devices. Gascon \textit{et al.} \cite{gascon2014continuous} collected freely typed samples from over 300 participants and developed a system that achieved a True Acceptance Rate (TAR) of $92\%$ at $1\%$ False Acceptance Rate (FAR) (an EER of about $10\%$). Their system utilized accelerometer, gyroscope, time, and orientation features. Each user typed an English pangram (sentence containing every letter of the alphabet) approximately 160 characters in length, and classification was performed by SVM. In other work, Kim and Kang \cite{kim2020freely} utilized microbehavioral features to obtain an EER below $0.05\%$ for 50 subjects with a single reference sample of approximately 200 keystrokes for both English and Korean input. The microbehavioral features consist of angular velocities along three axes when each key is pressed and released, as well as timing features and the coordinate of the touch event within each key. See \cite{teh2016survey} for a survey of keystroke biometrics on mobile devices.

Nowadays, with the proliferation of machine learning algorithms capable of analysing and learning human behaviors from large scale datasets, the performance of keystroke dynamics in the free-text scenario has been boosted. As an example, \cite{Ceker} proposes a combination of the existing digraphs method for feature extraction plus an SVM classifier to authenticate subjects. This approach achieves almost $0\%$ error rate using samples containing $500$ keystrokes. These results are very promising, even though it was evaluated using a small dataset with only $34$ subjects. In \cite{Deb} the authors employ an RNN within a Siamese architecture to authenticate subjects based on $8$ biometric modalities on smartphone devices. They achieved results in a free-text scenario of $81.61\%$ TAR at $0.1\%$ FAR using just $3$ second test windows with a dataset of $37$ subjects. 

\textcolor{black}{In \cite{cceker2017sensitivity}, the authors employed CNN (Convolutional Neural Network) with Gaussian data augmentation technique for fixed-text keystroke authentication over a population of $267$ subjects. Their results of $2.02\%$ EER in the best scenario suggest the combined benefit of CNN architectures and data augmentation for keystroke biometric systems. Finally, in \cite{Lu} the authors combined a CNN with a RNN architecture. They argued that adding a 1D convolutional layer at the top of the RNN architecture makes the model able to extract higher-level keystroke features that are processed by the following RNN layers. Their results tested with the SUNY Buffalo \cite{Buffalo2016} dataset showed a relative error reduction of $35\%$ (from $5.03\%$ to $2.67$\% EER) when employing the 1D convolutional layer with a population of $75$ users and keystrokes sequences of $30$ keys.}

Previous works in free-text keystroke dynamics have achieved promising results with up to several hundred subjects (see Table \ref{table:works}), but they have yet to scale beyond this limit and leverage emerging machine learning techniques that benefit from vast amounts of data. Here we take a step forward in this direction of machine learning-based free-text keystroke biometrics by using the largest datasets published to date with $199$ million keystrokes from $228$,$000$ subjects (considering both mobile and desktop datasets). We analyze to what extent deep learning models are able to scale in keystroke biometrics to recognize subjects at a large scale while attempting to minimize the amount of data per subject required for enrollment.

%\aythami{@Vinnie it would be great if you can add a comparison between mobile and desktop keystroke recognition}

%-------------------------------------------------------------------------

\section{Keystroke Datasets}
\label{aalto}
\textcolor{black}{The approaches proposed in this work were trained and evaluated using four public keystroke databases: 1) the Dhakal \textit{et al.} dataset \cite{Dhakal}; 2) the Palin \textit{et al.} dataset \cite{palin}; 3) the Clarkson II dataset \cite{Murphy}; and iv) the Buffalo dataset \cite{Buffalo2016}}. 

\textcolor{black}{The two Aalto University Datasets \cite{Dhakal}\cite{palin} were used for both training and evaluation. The Dhakal \textit{et al.} dataset \cite{Dhakal}} comprises more than $5$GB of keystroke data collected on desktop keyboards from $168$,$000$ participants;. The Palin \textit{et al.} dataset \cite{palin}, which comprises almost $4$GB of keystroke data collected on mobile devices from $260$,$000$ participants. The same data collection \textcolor{black}{based on controlled free-text \cite{wahab2021utilizing}} followed for both datasets. The acquisition task required subjects to memorize English sentences and then type them as quickly and accurate as they could. The English sentences were selected randomly from a set of $1$,$525$ examples taken from the Enron mobile email and Gigaword Newswire corpus. The example sentences contained a minimum of $3$ words and a maximum of $70$ characters. Note that the sentences typed by the participants could contain more than $70$ characters because each participant could forget or add new characters when typing. All participants in the Dhakal database completed $15$ sessions (i.e. one sentence for each session) on either a desktop or a laptop physical keyboard. However, in the Palin dataset the participants who finished at least $15$ sessions are only $23$\% ($60$,$000$ participants) out of $260$,$000$ participants that started the typing test. In this paper we will employ these $60$,$000$ subjects with their first 15 sessions in order to allow fair comparisons between both datasets.

For the data acquisition, the authors launched an online application that records the keystroke data from participants who visit their webpage and agree to complete the acquisition task (i.e. the data was collected in an uncontrolled environment). Press (keydown) and release (keyup) event timings were recorded in the browser with millisecond resolution using the JavaScript function \texttt{Date.now}. The authors also reported demographic statistics for both datasets: $72$\% of the participants from the Dhakal database took a typing course, $218$ countries were involved, and $85$\% of the them have English as native language, meanwhile only $31$\% of the participants from the Palin database took a typing course, $163$ countries were involved, and $68$\% of the them were English native speakers.

\textcolor{black}{The Clarkson II dataset \cite{Murphy} and Buffalo dataset \cite{Buffalo2016} were used to evaluate the generalization capacity of the trained models. The Clarkson II is composed by $103$ subjects typing in a desktop keyboard over a time span of $2.5$ years in a complete uncontrolled scenario (fully free-text). The data is not divided into sessions, so for each subject we split the entire raw keystroke data into keystroke sequences of length $150$. We only employ for testing the users that yield at least $15$ keystroke sequences according to this method. The Buffalo database is composed of $148$ subjects with $3$ sessions of keystroke data collected in desktop keyboards during $28$ days of time span. In each session, the subject completed two tasks: one task consisted of transcribing a prefixed text and the other task was based on answering free-text questions.}

\begin{figure}[t!]
\centering
\includegraphics[width=\columnwidth]{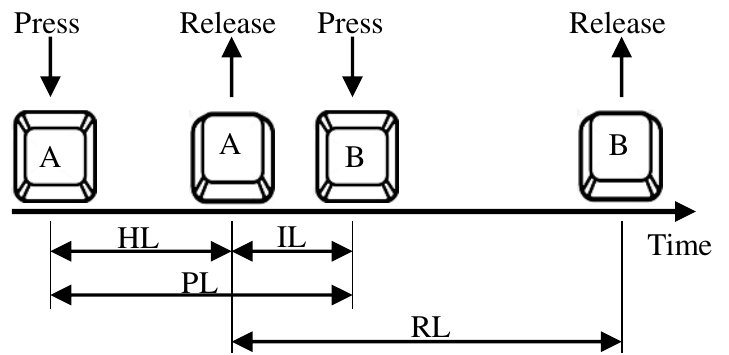}
\caption{Example of the 4 temporal features extracted between two consecutive keys: Hold Latency (HL), Inter-key Latency (IL), Press Latency (PL), and Release Latency (RL).}
\label{features}
\end{figure}

\section{System Description}
\label{system_description}
\subsection{Pre-processing and Feature Extraction}
The raw data captured in each session includes a time series with three dimensions: the keycodes, press times, and release times of the keystroke sequence. Timestamps are in UTC format with millisecond resolution, and the keycodes are integers between $0$ and $255$ according to the ASCII code.

We extract $4$ temporal features for each sequence (see Fig. \ref{features} for details): (i) Hold Latency (HL), the elapsed time between key press and release events; (ii) Inter-key Latency (IL), the elapsed time between releasing a key and pressing the next key; (iii) Press Latency (PL), the elapsed time between two consecutive press events; and (iv) Release Latency (RL), the elapsed time between two consecutive release events. These $4$ features are commonly used in both fixed-text and free-text keystroke systems \cite{Alsultan}. Finally, we include the keycodes as an additional feature.

The $5$ features are calculated for each keystroke in the sequence. Let $N$ be the length of the keystroke sequence, such that each sequence provided as input to the model is a time series with shape $N \times 5$ ($N$ keystrokes by $5$ features).
All feature values are normalized before being provided as input to the model. Normalization is important so that the activation values of neurons in the input layer of the network do not saturate (i.e. all close to $1$). The keycodes are normalized to between $0$ and $1$ by dividing each keycode by $255$, and the $4$ timing features are converted to seconds. This scales most timing features to between $0$ and $1$ as the average typing rate over the entire dataset is $5.1$ $\pm$ $2.1$ keys per second. Only latency features that occur either during very slow typing or long pauses exceed a value of $1$.
\subsection{TypeNet Architecture}
\begin{figure}[t!]
\centering
\includegraphics[width=0.6\columnwidth]{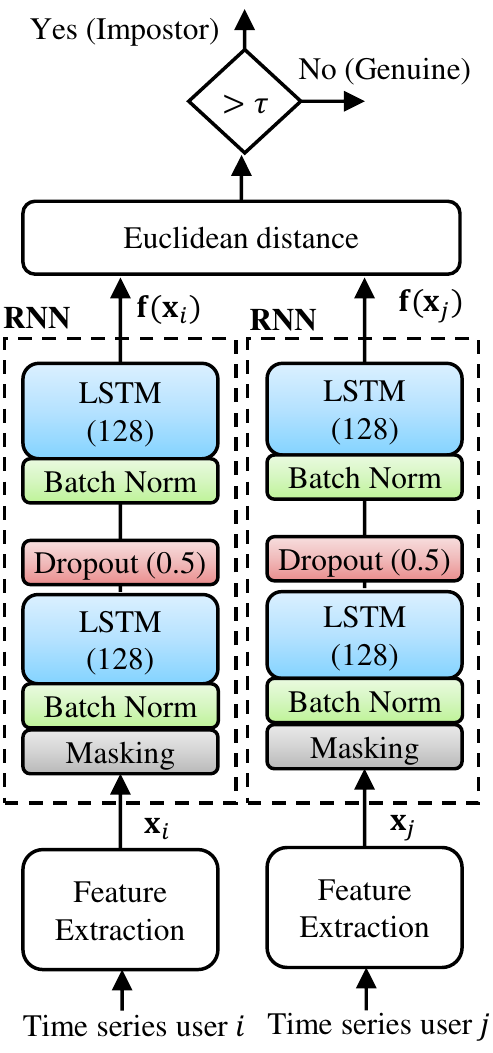}
\caption{\textcolor{Black}{Verification scheme based on the proposed architecture (TypeNet) for free-text keystroke sequences from two users $i$ and $j$. The input \textbf{x} is a time series with shape $M \times 5$ (keystrokes $\times$ keystroke features) and the output $\textbf{f}($\textbf{x}$)$ is an embedding vector with shape $1\times128$. $\tau$ is a decision threshold.}}
\label{LSTM}
\end{figure}
In keystroke dynamics, it is thought that idiosyncratic behaviors that enable authentication are characterized by the relationship between consecutive key press and release events (e.g. temporal patterns, typing rhythms, pauses, typing errors). In a free-text scenario, keystroke sequences between enrollment and testing may differ in both length and content. This reason motivates us to choose a Recurrent Neural Network as our keystroke authentication algorithm. RNNs have demonstrated to be one of the best algorithms to deal with temporal data (e.g. \cite{2020_TIFS_BioTouchPass2_Tolosana}, \cite{Tolosana1}) and are well suited for free-text keystroke sequences (e.g. \cite{Deb}, \cite{Lu}).

Our RNN architecture is depicted in Fig. \ref{LSTM}. It is composed of two Long Short-Term Memory (LSTM) layers of $128$ units (\textit{tanh} activation function). Between the LSTM layers, we perform batch normalization and dropout at a rate of $0.5$ to avoid overfitting. Additionally, each LSTM layer has a recurrent dropout rate of $0.2$. \textcolor{Black}{This RNN model is composed by $200$,$458$ trainable parameters. The proposed architecture is the result of several tests including different number of layers (from $1$ to $4$ layers) and units (from $8$ to $512$ units in powers of $2$). Our experiments showed that performance improvements were marginal for more than $2$ layers and more than $128$ units. We chosen this architecture to reduce the number of parameters and avoid a potential overfitting.} 

One constraint when training a RNN using standard backpropagation through time applied to a batch of sequences is that the number of elements in the time dimension (i.e. number of keystrokes) must be the same for all sequences. We set the size of the time dimension to $M$. In order to train the model with sequences of different lengths $N$ within a single batch, we truncate the end of the input sequence when $N>M$ and zero pad at the end when $N<M$, in both cases to the fixed size $M$. Error gradients are not computed for those zeros and do not contribute to the loss function at the output layer as a result of the masking layer shown in Fig.~\ref{LSTM}.

Finally, the output of the model $\textbf{f}($\textbf{x}$)$ is an array of size $1 \times 128$ that we will employ later as an embedding feature vector to recognize subjects.

\subsection{LSTM Training: Loss Functions}
Our goal is to build a keystroke biometric system capable of generalizing to new subjects not seen during model training, and therefore, having a competitive performance when it deploys to applications with thousands of users. Our RNN is trained only once on an independent set of subjects. This model then acts as a feature extractor that provides input to a distance-based recognition scheme. After training the RNN once, we will evaluate in the experimental section the recognition performance for a varying number of subjects and enrollment samples per subject.

We train our deep model with three different loss functions: \textit{Softmax loss}, which is widely used in classification tasks; \textit{Contrastive loss}, a loss for distance metric learning based on two samples \cite{ContrastiveLoss}; and \textit{Triplet loss}, a loss for metric learning based on three samples \cite{TripletLoss}. These are defined as follows.

\subsubsection{Softmax loss} 
\begin{figure*}[t!]
\centering
\includegraphics[width=0.95\textwidth]{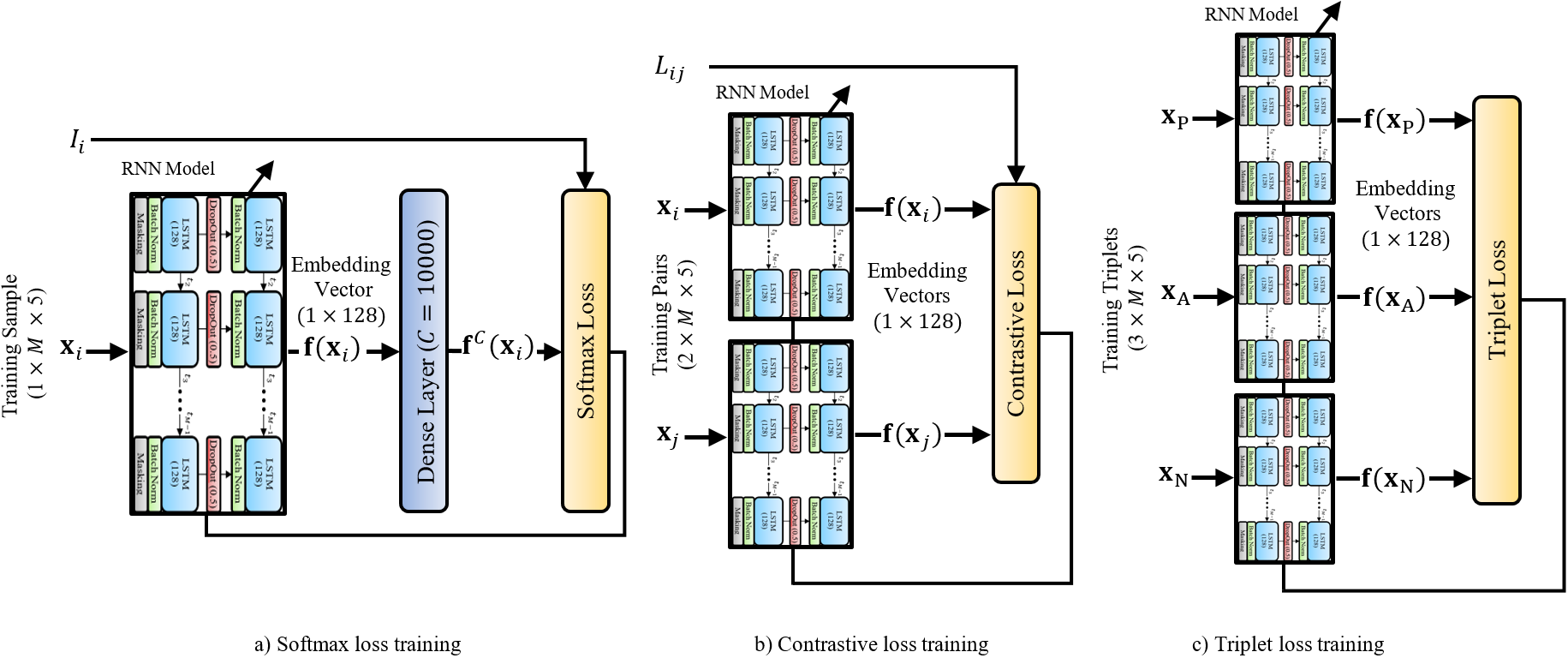}
\caption{Learning architecture for the different loss functions a) Softmax loss, b) Contrastive loss, and c) Triplet loss. The goal is to find the most discriminant embedding space $\textbf{f}(\textbf{x})$.}
\label{Siamese}
\end{figure*}

Let $\textbf{x}_{i}$ be a keystroke sequence of individual $I_i$, and let us introduce a dense layer after the embeddings described in the previous section aimed at classifying the individuals used for learning (see Fig.~\ref{Siamese}.a). The Softmax loss is applied as

\begin{equation}
\label{softmax}
     \mathcal{L}_{S}= -\log \left( \frac{e^{f^C_{I_i}(\textbf{x}_i)}}{\sum\limits_{c=1}^{C}e^{f^C_c(\textbf{x}_i)}}\right)
\end{equation}

\noindent where $C$ is the number of classes used for learning (i.e. identities), $\textbf{f}^C=[f^C_1,\ldots,f^C_c]$, and after learning all elements of $\textbf{f}^C$ will tend to 0 except $f^C_{I_i}(\textbf{x}_{i})$ that will tend to 1. \textcolor{Black}{Note that $C$ is a super-index to differentiate between $\textbf{f}$ and $\textbf{f}^C$, see Fig.~\ref{Siamese}(a).} Softmax is widely used in classification tasks because it provides good performance on closed-set problems. Nonetheless, Softmax does not optimize the margin between classes. Thus, the performance of this loss function usually decays for problems with high intra-class variance. In order to train the architecture proposed in Fig. \ref{LSTM}, we have added an output classification layer with $C$ units (see Fig. \ref{Siamese}.a). During the training phase, the model will learn discriminative information from the keystroke sequences and transform this information into an embedding space where the embedding vectors $\textbf{f}(\textbf{x})$ (the outputs of the model) will be close in case both keystroke inputs belong to the same subject (genuine pairs), and far in the opposite case (impostor pairs).

\subsubsection{Contrastive loss} Let $\textbf{x}_{i}$ and $\textbf{x}_{j}$ each be a keystroke sequence that together form a pair which is provided as input to the model. The Contrastive loss calculates the Euclidean distance between the model outputs,

\begin{equation}
\label{distance}
     d(\textbf{x}_{i},\textbf{x}_{j})= \left \| \textbf{f}(\textbf{x}_{i}) - \textbf{f}(\textbf{x}_{j})\right \|
\end{equation}

\noindent where $\textbf{f}(\textbf{x}_{i})$ and $\textbf{f}(\textbf{x}_{j})$ are the model outputs (embedding vectors) for the inputs $\textbf{x}_{i}$ and $\textbf{x}_{j}$, respectively. The model will learn to make this distance small (close to $0$) when the input pair is genuine and large (close to $\alpha$) for impostor pairs by computing the loss function $\mathcal{L}_{CL}$ defined as follows:

\begin{equation}
\label{loss}
     \mathcal{L}_{CL}= (1-L_{ij})\frac{d^2(\textbf{x}_{i},\textbf{x}_{j})}{2}+L_{ij}\frac{\max^2\left \{0, \alpha-d(\textbf{x}_{i},\textbf{x}_{j})\right \} }{2}
\end{equation}

\noindent where $L_{ij}$ is the label associated with each pair that is set to $0$ for genuine pairs and $1$ for impostor ones, and $\alpha \geq 0$ is the margin (the maximum margin between genuine and impostor distances). The Contrastive loss is trained using a Siamese architecture (see Fig. \ref{Siamese}.b) that minimizes the distance between embeddings vectors from the same class ($d(\textbf{x}_{i},\textbf{x}_{j})$ with $L_{ij}=0$), and maximizes it for embeddings from different class ($d(\textbf{x}_{i},\textbf{x}_{j})$ with $L_{ij}=1$).

\subsubsection{Triplet loss} The \textit{Triplet loss} function enables learning from positive and negative comparisons at the same time (note that the label $L_{ij}$ eliminates one of the distances for each pair in the Contrastive loss). A triplet is composed by three different samples from two different classes: Anchor (A) and Positive (P) are different keystroke sequences from the same subject, and Negative (N) is a keystroke sequence from a different subject. The Triplet loss function is defined as follows:

\begin{equation}
\label{TL_loss}
     \mathcal{L}_{TL}= \max \left \{0,d^2(\textbf{x}^{i}_\textrm{A},\textbf{x}^{i}_\textrm{P}) - d^2(\textbf{x}^{i}_\textrm{A},\textbf{x}^{j}_\textrm{N}) + \alpha \right \}
\end{equation}

\noindent where $\alpha$ is a margin between positive and negative pairs and $d$ is the Euclidean distance calculated with Eq. \ref{distance}. In comparison with Contrastive loss, Triplet loss is capable of learning intra- and inter-class structures in a unique operation (removing the label $L_{ij}$). The Triplet loss is trained using an extension of a Siamese architecture (see Fig. \ref{Siamese}.c) for three samples. This learning process minimizes the distance between embedding vectors from the same class ($d(\textbf{x}_{\textrm{A}},\textbf{x}_{\textrm{P}})$), and maximizes it for embeddings from different classes ($d(\textbf{x}_{\textrm{A}},\textbf{x}_{\textrm{N}})$).

\subsection{LSTM Training: Implementation Details}

We train three RNN versions (i.e. one for each loss function) for each input device: desktop and mobile, using the Dhakal and Palin databases, respectively. For the desktop scenario, we train the models using only the first $68$,$000$ subjects from the Dhakal dataset. The remaining $100$,$000$ subjects were employed only for model evaluation, so there is no data overlap between the two groups of subjects. This reflects an open-set authentication paradigm. For the Softmax function we train a model with $C= 10$,$000$ subjects due to GPU memory constraints, as the Softmax loss requires a very wide final layer with many classes. In this case, we used $15\times10$,$000=150$,$000$ keystroke sequences for training and the remaining $58$,$000$ subjects were discarded. For the Contrastive loss we generate genuine and impostor pairs using all the $15$ keystroke sequences available for each subject. This provides us with $15 \times 67$,$999 \times 15=15.3$ million impostor pair combinations and $15\times14/2=105$ genuine pair combinations for each subject. The pairs were chosen randomly in each training batch ensuring that the number of genuine and impostor pairs remains balanced ($512$ pairs in total in each batch including impostor and genuine pairs). Similarly, we randomly chose triplets for the Triplet loss training. 

The same protocol was employed for the mobile scenario but adjusting the amount of subjects employed to train and test. In order to have balanced subsets close to the desktop scenario, we divided by half the Palin database such that $30$,$000$ subjects were used to train the models, generating $15\times29$,$999\times15=6.75$ million impostor pair combinations and $15\times14/2=105$ genuine pair combinations for each subject. The other $30$,$000$ subjects were used to test the mobile TypeNet models. Once again $10$,$000$ subjects were used to train the mobile TypeNet model with Softmax loss. 

Regarding the hyper-parameters employed during training, the best results for both models were achieved with a learning rate of $0.05$, Adam optimizer with $\beta_{1} = 0.9$, $\beta_{2} = 0.999$ and $\epsilon = 10^{-8}$, and the margin set to $\alpha = 1.5$. The models were trained for $200$ epochs with 150 batches per epoch and $512$ sequences in each batch. The models were built in \texttt{Keras-Tensorflow}.

\section{Experimental Protocol}
\label{experimental_protocol}
\subsection{Authentication Protocol}
We authenticate subjects by comparing gallery samples $\textbf{x}_{i,g}$ belonging to the subject $i$ in the test set to a query sample  $\textbf{x}_{j,q}$ from either the same subject (genuine match $i=j$) or another subject (impostor match $i \ne j$). The test score is computed by averaging the Euclidean distances between each gallery embedding vector $\textbf{f}(\textbf{x}_{i,g})$ and the query embedding vector $\textbf{f}(\textbf{x}_{j,q})$ as follows:
\begin{equation}
\label{score}
     \textit{s}_{i,j}^q= \frac{1}{G}\sum_{g=1}^{G} ||\textbf{f}(\textbf{x}_{i,g})-\textbf{f}(\textbf{x}_{j,q})||
\end{equation}
where $G$ is the number of sequences in the gallery (i.e. the number of enrollment samples) and $q$ is the query sample of subject $j$. Taking into account that each subject has a total of $15$ sequences, we retain $5$ sequences per subject as the test set (i.e. each subject has $5$ genuine test scores) and let $G$ vary between $1 \leq G \leq 10$ in order to evaluate the performance as a function of the number of enrollment sequences.

To generate impostor scores, for each enrolled subject we choose one test sample from each remaining subject. We define $k$ as the number of enrolled subjects. In our experiments, we vary $k$ in the range $100 \leq k \leq K$, where $K = 100$,$000$ for the desktop TypeNet models and $K = 30$,$000$ for the mobile TypeNet. Therefore each subject has $5$ genuine scores and $k-1$ impostor scores. Note that we have more impostor scores than genuine ones, a common scenario in keystroke dynamics authentication. The results reported in the next section are computed in terms of Equal Error Rate (EER), which is the value where False Acceptance Rate (FAR, proportion of impostors classified as genuine) and False Rejection Rate (FRR, proportion of genuine subjects classified as impostors) are equal. The error rates are calculated for each subject and then averaged over all $k$ subjects \cite{2014_IWSB_Aythami_Keystroking}.

\subsection{Identification Protocol}
Identification scenarios are common in forensics applications, where the final decision is based on a bag of evidences and the biometric recognition technology can be used to provide a list of candidates, referred to as background set $\mathfrak{B}$ in this work. The Rank-1 identification rate reveals the performance to unequivocally identifying the target subject among all the subjects in the background set. Rank-$n$ represents the accuracy if we consider a ranked list of $n$ profiles from which the result is then manually or automatically determined based on additional evidence \cite{2018_INFFUS_MCSreview2_Fierrez}. 

The $15$ sequences from the $k$ test subjects in the database were divided into two groups: Gallery ($10$ sequences) and Query ($5$ sequences). We evaluate the identification rate by comparing the Query set of samples $\textbf{x}_{j,q}^{\textrm{Q}}$, with $q=1,...,5$ belonging to the test subject $j$ against the Background Gallery set $\textbf{x}_{i,g}^{\textrm{G}}$, with $g=1,...,10$  belonging to all background subjects. The distance was computed by averaging the Euclidean distances $||\cdot||$ between each gallery embedding vector $\textbf{f}(\textbf{x}_{i,g}^{\textrm{G}})$ and each query embedding vector $\textbf{f}(\textbf{x}_{j,q}^{\textrm{Q}})$ as follows:
\begin{equation}
\label{score_identification}
     s_{i,j}^{Q}= \frac{1}{10 \times 5}\sum_{g=1}^{10}\sum_{q=1}^{5} ||\textbf{f}(\textbf{x}_{i,g}^{\textrm{G}})-\textbf{f}(\textbf{x}_{j,q}^{\textrm{Q}})||
\end{equation}

We then identify a query set (i.e. subject $j=J$ is the same gallery person $i=I$) as follows:

\begin{equation}
\label{mindistance}
     I = \arg\min_i s_{i,J}^{Q}
\end{equation}

The results reported in the next section are computed in terms of Rank-$n$ accuracy. A Rank-$1$ means that $d_{i,J}<d_{I,J}$ for any $i \neq I$, while a Rank-$n$ means that instead of selecting a single gallery profile, we select $n$ profiles starting with $i=I$ by increasing distance $d_{i,J}$. In forensic scenarios, it is traditional to use Rank-20, Rank-50, or Rank-100 in order to generate a short list of potential candidates that are finally identified by considering other evidence.

\section{Experiments and Results}
\label{experiments_results}
%-------------------------------------------------------------------------
\setlength{\tabcolsep}{7pt}
\renewcommand{\arraystretch}{2.0}
\begin{table*}
\begin{center}
\begin{tabular}{cc|c|c|c|c|c|}
\cline{3-7}
\multicolumn{2}{c}{} &\multicolumn{5}{|c|}{\textbf{\#enrollment sequences per subject $G$}} \\ 
\cline{3-7} 
\multicolumn{2}{c|}{\multirow{-2}{*}{}} & \textbf{1} & \textbf{2} & \textbf{5} & \textbf{7} & \textbf{10} \\ \hline
\multicolumn{1}{|c|}{} & \textbf{30}  & 17.2/10.7/8.6  & 14.1/9.0/6.4  & 13.3/7.3/4.6 & 12.7/6.8/4.1   &  11.5/3.3/3.7\\
\cline{2-7} 
\multicolumn{1}{|c|}{}  & \textbf{50}  
& 16.8/8.2/5.4          & 13.1/6.7/3.6           & 10.8/5.4/2.2         & 9.2/4.8/1.8                       & 8.8/4.3/1.6                               \\ \cline{2-7} 
\multicolumn{1}{|c|}{}                         & \textbf{70}  
& 14.1/7.7/4.5      & 10.4/6.2/2.8        & 7.5/4.8/1.7      &  6.7/4.3/1.4               & 6.0/3.9/1.2                               \\ \cline{2-7} 
\multicolumn{1}{|c|}{}                         & \textbf{100} 
& 13.8/7.7/4.2           & 10.1/6.0/2.7                & 7.4/4.7/1.6                    & 6.4/4.3/1.4         & 5.7/3.9/1.2                              \\ \cline{2-7} 
\multicolumn{1}{|c|}{\multirow{-5}{*}{\rotatebox{90}{\textbf{\#keys per sequence $M$}}}} & \textbf{150}
& 13.8/7.7/4.1          & 10.1/6.0/2.7                    & 7.4/4.7/1.6    & 6.5/4.3/1.4               & 5.8/3.8/1.2                                \\ \hline
\end{tabular}
\end{center}
\caption{Equal Error Rates ($\%$) achieved in \textbf{desktop}  scenario using Softmax/Contrastive/Triplet loss for different values of the parameters $M$ (sequence length) and $G$ (number of enrollment sequences per subject).}
\label{table:performance_dektop}
\end{table*}

\setlength{\tabcolsep}{7pt}
\renewcommand{\arraystretch}{2.0}
\begin{table*}
\begin{center}
\begin{tabular}{cc|c|c|c|c|c|}
\cline{3-7}
\multicolumn{2}{c}{} &\multicolumn{5}{|c|}{\textbf{\#enrollment sequences per subject $G$}} \\ 
\cline{3-7} 
\multicolumn{2}{c|}{\multirow{-2}{*}{}} & \textbf{1} & \textbf{2} & \textbf{5} & \textbf{7} & \textbf{10} \\ \hline
\multicolumn{1}{|c|}{} & \textbf{30}  & 17.7/15.7/14.2   & 16.0/14.1/12.5   & 15.2/13.0/11.3  & 14.9/12.6/10.9    &  14.5/12.1/10.5 \\
\cline{2-7} 
\multicolumn{1}{|c|}{}  & \textbf{50}  
& 17.2/14.6/12.6          & 15.4/13.1/10.7           & 13.8/12.1/9.2          & 13.4/11.5/8.5                       & 12.7/11.0/8.0                                \\ \cline{2-7} 
\multicolumn{1}{|c|}{}                         & \textbf{70}  
& 17.8/13.8/11.3      & 15.5/12.4/9.5        & 13.5/11.2/7.8       &  13.0/10.7/7.2                & 12.1/10.4/6.8                                \\ \cline{2-7} 
\multicolumn{1}{|c|}{}                         & \textbf{100} 
& 18.4/13.6/10.7            & 15.8/12.3/8.9                 & 13.6/10.9/7.3                    & 13.0/10.4/6.6          & 12.3/10.0/6.3                               \\ \cline{2-7} 
\multicolumn{1}{|c|}{\multirow{-5}{*}{\rotatebox{90}{\textbf{\#keys per sequence $M$}}}} & \textbf{150}
& 18.4/13.7/10.7           & 15.9/12.3/8.8                     & 13.7/10.8/7.3     & 13.0/10.4/6.6                & 12.3/10.0/6.3                                \\ \hline
\end{tabular}
\end{center}
\caption{Equal Error Rates ($\%$) achieved in \textbf{mobile} scenario using Softmax/Contrastive/Triplet loss for different values of the parameters $M$ (sequence length) and $G$ (number of enrollment sequences per subject).}
\label{table:performance_mobile}
\end{table*}

%-------------------------------------------------------------------------
\subsection{Authentication Results}
As commented in the related works section, one key factor when analyzing the performance of a free-text keystroke authentication algorithm is the amount of keystroke data per subject employed for enrollment \cite{Huang_2015}. In this work, we study this factor with two variables: the keystroke sequence length $M$ and the number of gallery sequences used for enrollment $G$.

Our first experiment reveals to what extent $M$ and $G$ affect the authentication performance of our TypeNet models. Note that the input to our models has a fixed size of $M$ after the masking process shown in Fig.~\ref{LSTM}. For this experiment, we set $k = 1$,$000$ (where $k$ is the number of enrolled subjects).
Tables \ref{table:performance_dektop} and \ref{table:performance_mobile} summarize the error rates in both desktop and mobile scenarios respectively, achieved by the TypeNet models for the different values of sequence length $M$ and enrollment sequences per subject $G$. 

In the desktop scenario (Table \ref{table:performance_dektop}) we  observe that for sequences longer than $M = 70$ there is no significant improvement in performance. Adding three times more key events (from $M = 50$ to $M = 150$) lowers the EER by only $0.7\%$ in average for all values of $G$. However, adding more sequences to the gallery shows greater improvements with about $50\%$ relative error reduction when going from $1$ to $10$ sequences independent of $M$. Comparing among the different loss functions, the best results are always achieved by the model trained with Triplet loss for $M = 70$ and $G = 10$ with an error rate of $1.2\%$ \textcolor{black}{(with a standard deviation of $ \sigma \leq 4.1\%$)}, followed by the Contrastive loss function with an error rate of $3.9\%$; the worst results are achieved with the Softmax loss function ($6.0\%$). For one-shot authentication ($G = 1$), our approach has an error rate of $4.5\%$ using sequences of $70$ keystrokes. 

\begin{figure*}[t!]
\centering
\includegraphics[scale=0.35]{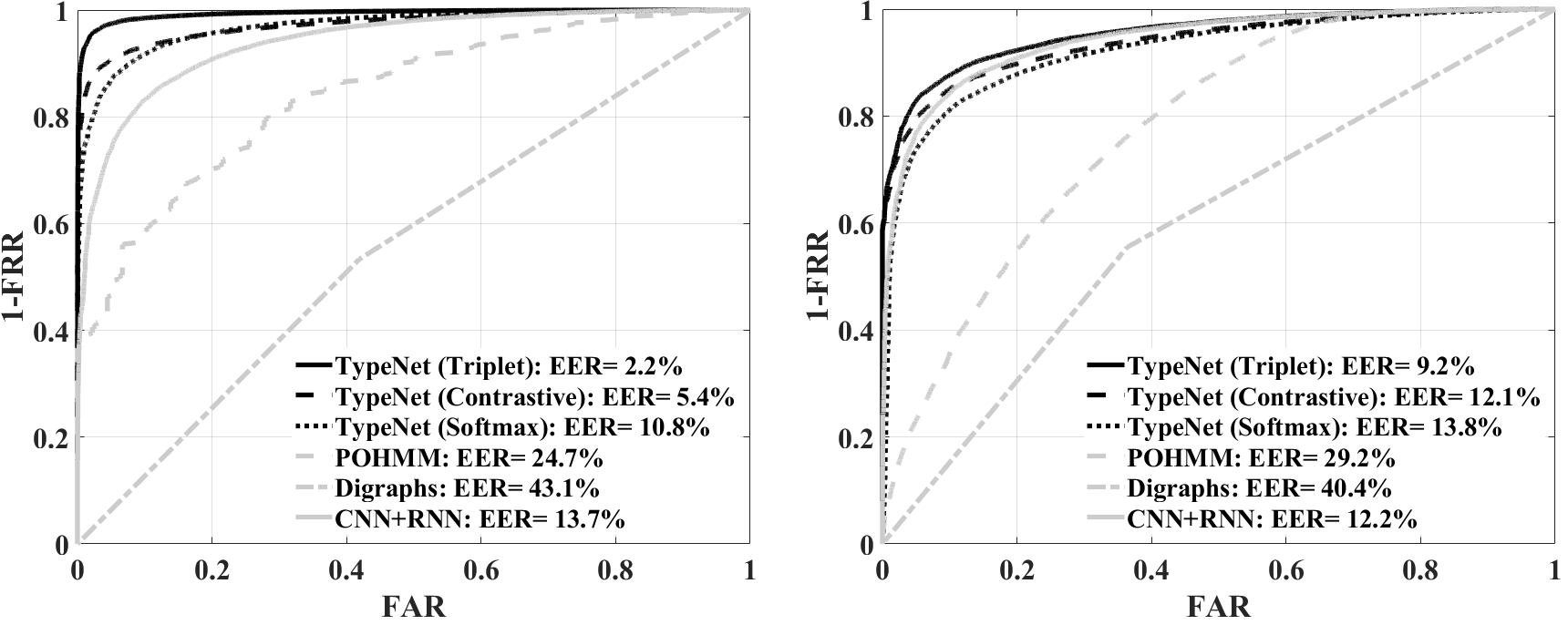}
\caption{\textcolor{black}{ROC comparisons in free-text biometric authentication for desktop (left) and mobile (right) scenarios between the three proposed TypeNet models and three state-of-the-art approaches: POHMM \textcolor{Black}{(Partially Observable Hidden Markov Model)} from \cite{Monaco}, digraphs/SVM from \cite{Ceker}, and CNN+RNN \textcolor{Black}{(Convolutional Neuronal Network + Recurrent Neuronal Network)} model from \cite{Lu}. $M = 50$ keystrokes per sequence, $G = 5$ enrollment sequences per subject, and $k = 1$,$000$ test subjects.}}
\label{ROCs}
\end{figure*}

Similar trends are observed in the mobile scenario (Table \ref{table:performance_mobile}) compared to the desktop scenario (Table \ref{table:performance_dektop}). First, increasing sequence length beyond $M = 70$ keystrokes does not significantly improve performance, but there is a significant improvement when increasing the number of sequences per subject. The best results are achieved for $M = 100$ and $G = 10$ with an error rate of $6.3\%$ by the model trained with triplet loss \textcolor{black}{(with a standard deviation of $\sigma \leq 9.2\%$)}, followed again by the contrastive loss ($10.0\%$), and softmax ($12.3\%$). For one-shot authentication ($G = 1$), the performance of the triplet model decays up to $10.7\%$ EER using sequences of $M = 100$ keystrokes. 

Comparing the performance achieved by the three TypeNet models between mobile and desktop scenarios, we observe that in all cases the results achieved in the desktop scenario are significantly better to those achieved in the mobile scenario. These results are consistent with prior work that has obtained lower performance on mobile devices when only timing features are utilized \cite{Buschek,teh2016survey,Banovic}.

Next, we compare TypeNet with our implementation of three state-of-the-art algorithms for free-text keystroke authentication: a statistical sequence model, the POHMM (Partially Observable Hidden Markov Model) from \cite{Monaco}, an algorithm based on digraphs and SVM from \cite{Ceker}, \textcolor{black}{and a deep model based on the combination of CNN and RNN architectures introduced in \cite{Lu}}. To allow fair comparisons, all approaches are trained and tested with the same data and experimental protocol: $G = 5$ enrollment sequences per subject, $M = 50$ keystrokes per sequence, $k = 1$,$000$ test subjects. \textcolor{black}{The CNN+RNN architecture proposed in \cite{Lu} was trained following  the  same  protocol employed  with  the  TypeNet model.}

In Fig. \ref{ROCs} we plot the error rates of the four approaches (i.e. Digraphs, POHMM, CNN+RNN, and TypeNet) trained and tested on both desktop (left) and mobile (right) datasets. The TypeNet models outperform previous state-of-the-art free-text algorithms in both mobile and desktop scenarios with this experimental protocol, where the amount of enrollment data is reduced ($5 \times M=250$ training keystrokes in comparison to more than $10$,$000$ in related works, see Section~\ref{related_works}). This can largely be attributed to the rich embedding feature vector produced by TypeNet, which minimizes the amount of data needed for enrollment. The SVM generally requires a large number of training sequences per subject ($\sim$$100$), whereas in this experiment we have only $5$ training sequences per subject. We hypothesize that the lack of training samples contributes to the poor performance (near chance accuracy) of the Digraphs system based on SVMs. \textcolor{black}{Finally, the results achieved by the model based on CNN+RNN are the closest to those achieved by the TypeNet models. The deep learning architectures clearly outperform traditional approaches. However, the performance of TypeNet is significantly better than the performance achieved by the architecture proposed in \cite{Lu}, especially for the desktop scenario.}

\subsection{Authentication: Varying Number of Subjects}
In this experiment, we evaluate to what extent our best TypeNet models (those trained with triplet loss) are able to generalize without performance decay. For this, we scale the number of enrolled subjects $k$ from $100$ to $K$ (with $K = 100$,$000$ for desktop and $K = 30$,$000$ for mobile). For each subject we have $5$ genuine test scores and $k - 1$ impostor scores, one against each other test subject. The models used for this experiment are the same trained in previous the section ($68$,$000$ independent subjects included in the training phase for desktop and $30$,$000$ for mobile). 
\begin{figure}[t!]
\centering
\includegraphics[width=\columnwidth]{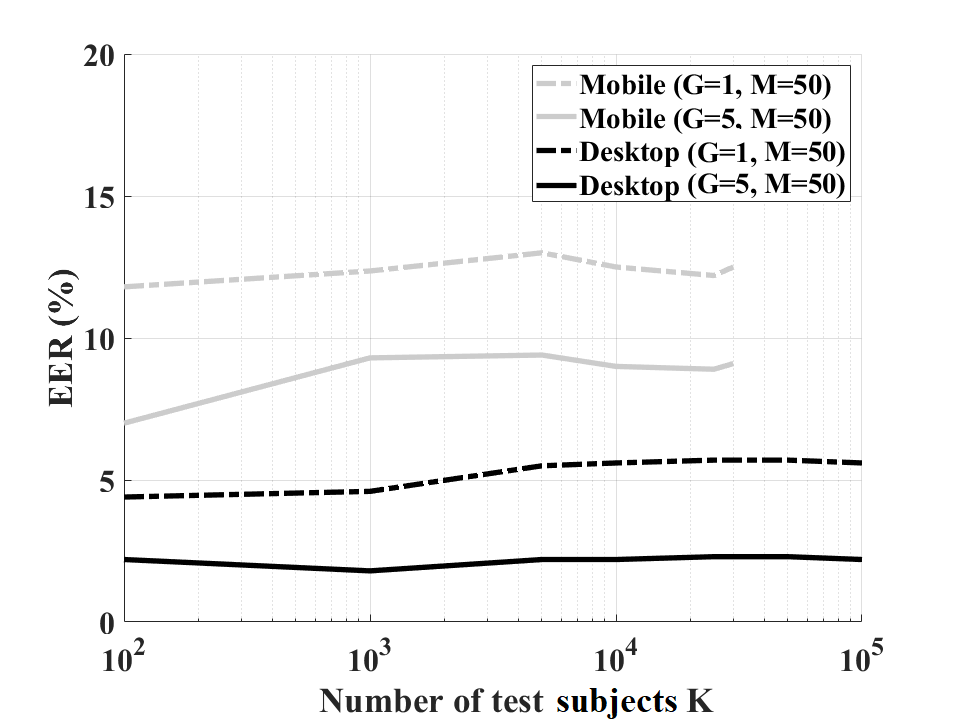}
\caption{EER (\%) of our proposed TypeNet models when scaling up the number of test subjects $k$ in one-shot ($G = 1$ enrollment sequences per subject) and 5-shot ($G = 5$) authentication cases. $M = 50$ keystrokes per sequence.}
\label{EERs}
\end{figure}

Fig. \ref{EERs} shows the authentication results for one-shot enrollment ($G = 1$ enrollment sequences, $M = 50$ keystrokes per sequence) and the case ($G = 5$, $M = 50$) for different values of $k$. For the desktop devices, we can observe that in both cases there is a slight performance decay when we scale from $1$,$000$ to $10$,$000$ test subjects, which is more pronounced in the one-shot case. However, for a large number of subjects ($k \geq 10$,$000$), the error rates do not appear to demonstrate continued growth. For the mobile scenario, the results when scaling from $100$ to $1$,$000$ test subjects show a similar tendency compared to the desktop scenario with a slightly greater performance decay. However, we can observe an error rate reduction when we continue scaling the number of test subjects up to $30$,$000$. In all cases the variation of the performance across the number of test subjects is less than $2.5\%$ EER. These results demonstrate the potential of the RNN architecture in TypeNet to authenticate subjects at large scale in free-text keystroke dynamics. We note that in the mobile scenario, we have utilized only timing features; prior work has found that greater performance may be achieved by incorporating additional sensor features \cite{kim2020freely}.

\subsection{Authentication: Cross-database Interoperability}
In this experiment we measure the cross-device interoperability between the best TypeNet models trained with the triplet loss. We also study the capacity of both desktop and mobile TypeNet models to generalize to other input devices and state-of-the-art databases. For this, we test both models with a different keystroke dataset than the one employed in their training. Additionally, for this experiment we train a third TypeNet model called Mixture-TypeNet with triplet loss using keystroke sequences from both datasets (half of the training batch for each dataset) but keeping the same train/test subject division as the other TypeNet models to allow fair comparisons. To be consistent with the other experiments we keep the same experimental protocol: $G = 5$ enrollment sequences per subject, $M = 50$ keystrokes per sequence, $k = 1$,$000$ test subjects.

\setlength{\tabcolsep}{9pt}

\begin{table}[t]
  \begin{center}
\begin{tabular}{|lc|c|c|c|}
\cline{3-5}
\multicolumn{2}{l}{} &
  \multicolumn{3}{|c|}{\textbf{TypeNet model}} \\ \cline{3-5} 
\multicolumn{2}{l|}{\multirow{-2}{*}{}} & \textbf{Desktop} & \textbf{Mobile} & \textbf{Mixture} \\ \hline
\multicolumn{1}{|c|}{} &
  \textbf{Aalto Desktop} &
  2.2 & 21.4 & 17.9\\ \cline{2-5} 
\multicolumn{1}{|c|}{} &
  \textbf{Aalto Mobile} &
  13.7 & 9.2 & 12.6\\ \cline{2-5}
  \multicolumn{1}{|c|}{} &
  \textbf{\textcolor{black}{Buffalo (Free) }} &
  7.6 & 33.2 & 22.1\\ \cline{2-5} 
\multicolumn{1}{|c|}{\multirow{-4}{*}{\rotatebox{90}{\textbf{ Test dataset }}}} &
  \textbf{\textcolor{black}{Buffalo (Transc)}} &
9.5 & 32.8 & 23.1 \\ \cline{2-5}
\multicolumn{1}{|c|}{} &
  \textbf{\textcolor{black}{Clarkson II}} &
  26.8 & 36.6 & 35.8\\ \cline{2-5}
  \multicolumn{1}{|c|}{} &
  \textbf{\textcolor{black}{Clarkson II*}} &
  17.2 & 33.0 & 30.4\\ 
  \hline
\end{tabular}
\end{center}
\caption{\textcolor{black}{Equal Error Rates (\%) achieved in the cross-database scenario for the three TypeNet models (Desktop, Mobile, and Mixture) when testing on Aalto Desktop \cite{Dhakal}, Aalto Mobile \cite{palin}, Clarkson II \cite{ayotte2020fast}, and Buffalo \cite{Buffalo2016} dataset. Buffalo (Free) $=$ free text, Buffalo (Transc) $=$ transcripted text. *Experiment using all the data available per subject.}}
\label{table:cross_sensor}

\end{table}

First of all, Table \ref{table:cross_sensor} shows the error rates achieved for the three TypeNet models when we test with or the three Type-Net models when we test with desktop (Dhakal) and mobile (Palin) datasets (named in this section Aalto Desktop and Aalto Mobile respectively for a better comparison with other databases). First, We can observe that error rates increase significantly in the cross-device scenario for both desktop and mobile TypeNet models. This performance decay is alleviated by the Mixture-TypeNet model, which still performs much worse than the other two models trained and tested in the same-sensor scenario. These results suggest that multiple device-specific models may be superior to a single model when dealing with input from different device types. This would require device type detection in order to pass the enrollment and test samples to the correct model \cite{alonso10qualityBasedSensorInteroperability}. 

\textcolor{black}{Secondly, we test the generalization capacity of the three TypeNet models with two public free-text keystroke databases: the Clarkson II dataset collected in \cite{Murphy} and the Buffalo dataset collected in \cite{Buffalo2016}. Table \ref{table:cross_sensor} presents the performance of the proposed approaches over the Clarkson II database and Buffalo database in transcribed (Transc) and free-text (Free) scenarios. Note that the models were trained and evaluated with different databases. This experiment is aimed to explore the generalization capacity between various data collection environments. Due to the number of subjects in both Clarkson II and Buffalo databases, which is much fewer than those present in the Aalto datasets, we modified the experimental protocol. For Clarkson II we employed $k = 91$ (the number of subjects for which we could extract at least $15$ samples of $150$ keys), $G = 5$ enrollment sequences per subject, $M = 50$ keystrokes per sequence. For the Buffalo database we employed $k = 147$, $G = 2$ enrollment sequences per subject (as we only have three sessions per subject, we employ two for gallery and one for query), and $M = 50$ keystrokes per sequence.} 

\textcolor{black}{The last three rows of Table \ref{table:cross_sensor} show the results achieved when testing with both Clarkson II and Buffalo databases. The performance of the Desktop version of TypeNet remained competitive for the Bufallo dataset even when we only employed $G= 2$ gallery samples per subject. Nonetheless, there is a large increase of the error rates for Clarkson II database. This drop of performance might be caused by the uncontrolled acquisition of the Clarkson II database over a long time period (i.e. two years) and the fully free-text typing behavior. However, when we employ all keystroke data available in the database per subject for testing (i.e. $G=10$ and $M=150$) the error rate drops up to $17.2\%$. Note that the benchmark published in \cite{Murphy} achieved EERs around $10\%$ training and testing with the same database. The results obtained by the owner of the database demonstrate the \textcolor{Black}{uncontrolled} conditions of this database. We want to highlight that the TypeNet models were not retrained with any kind of keystroke data from Clarkson II or Buffalo databases, these databases were employed only for testing. These results suggest that re-training is necessary to improve the performance of the proposed models, especially for the Clackson II database. On the other hand, the performance achieved by the Mobile and Mixed versions of TypeNet was very poor with EERs greater than $20\%$. Both databases were acquired with desktop keyboards and these results indicates the importance of the device in the generalization capacity of the models.}

\subsection{Identification based on Keystroke Dynamics}

Table \ref{table:table_identification} presents the identification accuracy for a background of $\mathfrak{B}=1$,$000$ subjects, $k = 10$,$000$ test subjects, $G = 10$ gallery sequences per subject, and $M = 50$ keystrokes per sequence. The accuracy obtained for an identification scenario is much lower than the accuracy reported for authentication. In general, the results suggest that keystroke identification enables a $90\%$ size reduction of the candidate list while maintaining almost $100\%$ accuracy (i.e. $100\%$ rank-100 accuracy with $1$,$000$ subjects). However, the results show the superior performance of the triplet loss function and significantly better performance compared to traditional keystroke approaches \cite{Ceker, Monaco, Lu}. While traditional approaches are not suitable for large-scale free text keystroke applications, the results obtained by TypeNet demonstrate its usefulness in many applications. 

The number of background profiles can be further reduced if auxiliary data is available to realize a pre-screening of the initial list of gallery profiles (e.g. country, language). The Aalto University Dataset contains auxiliary data including age, country, gender, keyboard type (desktop vs laptop), among others. Table \ref{table:table_acc} shows also subject identification accuracy over the $1$,$000$ subjects with a pre-screening by country (i.e. contents generated in a country different to the country of the target subject are removed from the background set). The results show that pre-screening based on a unique attribute is enough to largely improve the identification rate: Rank-1 identification with pre-screening ranges between $5.5\%$ to $84.0\%$, while the Rank-100 ranges between $42.2\%$ to $100\%$. These results demonstrate the potential of keystroke dynamics for large-scale identification when auxiliary information is available.        

\begin{table}[t]
\renewcommand{\arraystretch}{1.5}
  \begin{center}
    \begin{tabular}{|l|c|c|c|c|} 
     % \toprule
      \hline
      \multirow{2}{*}{\textbf{Method}} & \multirow{2}{*}{\textbf{Scenario}} & \multicolumn{3}{c|}{\textbf{Rank}} \\
      \cline{3-5}
      & &  \textbf{1} &  \textbf{50} &  \textbf{100}  \\
     \hline
     \hline
      %\midrule % <-- Midrule here
      %\Tspace
      Digraph \cite{Ceker} & D & $0.1$ & $9.5$ & $15.2$ \\ % <-- Content of first column omitted.
      \hline
      
      %\midrule % <-- Midrule here
      %\Tspace
      POHMM \cite{Monaco} & D & $6.1$ & $48.4$  & $63.4$  \\
      \hline
      CNN+RNN \cite{Lu} & D & $44.2$ & $95.5$  & $98.2$  \\
      \hline
      TypeNet (\textit{softmax}) & D & $47.5$ & $96.3$  & $98.7$  \\
      \hline
      TypeNet (\textit{contrastive}) & D & $29.4$ & $97.2$  & $99.3$ \\
      \hline
      TypeNet (\textit{triplet}) & D & $67.4$ & $99.8$  & $99.9$ \\
      \hline
      \hline
      Digraph \cite{Ceker} & M& $0.0$ & $8.5$ & $14.4$ \\ % <-- Content of 
      \hline
      POHMM \cite{Monaco} & M & $6.5$ & $41.8$  & $53.7$  \\
      \hline
      CNN+RNN \cite{Lu} & M & $24.5$ & $86.3$  & $90.5$  \\
      \hline
      TypeNet (\textit{softmax}) & M & $23.5$ & $82.6$  & $91.4$  \\
      \hline
      TypeNet (\textit{contrastive}) & M & $19.0$ & $80.4$  & $89.8$ \\
      \hline
      TypeNet (\textit{triplet}) & M & $25.5$ & $87.5$  & $94.2$ \\
      \hline
      
      %Ours (VGG-Face) & 10.03(21\%) & 1.98 (43\%) & 1.69 (61\%)\\ % <-- Content of first column omitted.
      %\bottomrule % <-- Bottomrule here
    \end{tabular}
  \end{center}
      \caption{Identification accuracy (Rank-$n$ in \%) for a background size $\mathfrak{B}=1$,$000$. Scenario: D = Desktop, M = Mobile.}
\label{table:table_identification}
\end{table}

\begin{table}[t]
\renewcommand{\arraystretch}{1.5}
  \begin{center}
    \begin{tabular}{|l|c|c|c|c|} 
     % \toprule
      \hline
      \multirow{2}{*}{\textbf{Method}} & \multirow{2}{*}{\textbf{Scenario}} & \multicolumn{3}{c|}{\textbf{Rank}} \\
      \cline{3-5}
      &  & \textbf{1} & \textbf{50} & \textbf{100}  \\
     \hline
      %\midrule % <-- Midrule here
      %\Tspace
      Digraph \cite{Ceker} & D & $5.5$ & $37.6$ & $42.2$ \\ % <-- Content of first column omitted.
      \hline
      %\midrule % <-- Midrule here
      %\Tspace
      POHMM \cite{Monaco} & D & $21.8$ & $78.3$  & $89.7$  \\
      \hline
      CNN+RNN \cite{Lu} & D & $65.1$ & $99.1$  & $99.7$  \\
      \hline
      TypeNet (\textit{softmax}) & D & $68.3$ & $99.39$  & $99.9$  \\
      \hline
      TypeNet (\textit{contrastive}) & D & $56.3$ & $99.7$  & $99.9$ \\
      %\midrule % <-- Midrule here
      %\Tspace
      \hline
      TypeNet (\textit{triplet}) & D & $84.0$ & $99.9$  & $100$ \\
      \hline
      %Ours (VGG-Face) & 10.03(21\%) & 1.98 (43\%) & 1.69 (61\%)\\ % <-- Content of first column omitted.
      %\bottomrule % <-- Bottomrule here
    \end{tabular}
  \end{center}
      \caption{Identification accuracy (Rank-$n$ in \%) for a background size $\mathfrak{B}=1$,$000$ and pre-screening based on the location of the typist. Scenario: D = Desktop. There is not metadata related to the mobile scenario.}
\label{table:table_acc}
\end{table}

\subsection{Input Text Dependency in TypeNet Models}
For the last experiment, we examine the effect of the text typed (i.e. the keycodes employed as input feature in the TypeNet models) on the distances between embedding vectors and how this may affect the model performance. The main drawback when using the keycode as an input feature to free-text keystroke algorithms is that the model could potentially learn text-based features (e.g. orthography, linguistic expressions, typing styles) rather than keystroke dynamics (e.g., typing speed and style) features. To analyze this phenomenon, we first introduce the Levenshtein distance (commonly referred as \textit{Edit distance}) proposed in \cite{editdistance}. The Levenshtein distance $d_L$ measures the distance between two words as the minimum number of single-character edits (insertions, deletions or substitutions) required to change one word into another. As an example, the Levenshtein distance between \textit{``kitten"} and \textit{``sitting"} is $d_L = 3$, because we need to substitute \textit{``s"} for \textit{``k"}, substitute \textit{``i"} for \textit{``e"}, and insert \textit{``g"} at the end (three editions in total). With the Levenshtein distance metric we can measure the similarity of two keystroke sequences in terms of keys pressed and analyze whether TypeNet models could be learning linguistic expressions to recognize subjects. This would be revealed by a high correlation between Levenshtein distance $d_L$ and the Euclidean distance of test scores $d_E$.

\begin{figure*}[t!]
\centering
\includegraphics[width=0.95\textwidth]{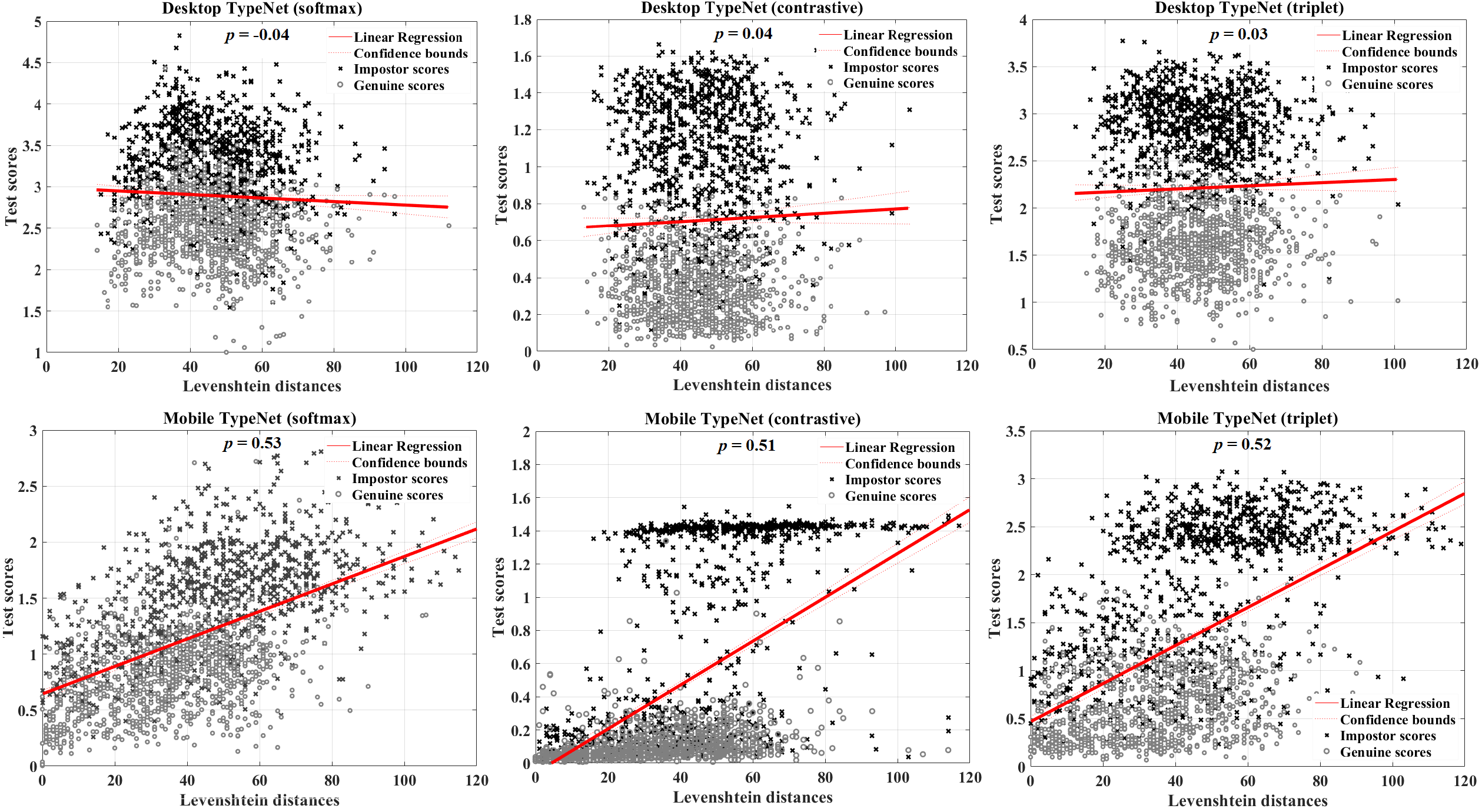}
\caption{Levenshtein distances vs. test scores in desktop (up) and mobile (down) scenarios for the three TypeNet models. For qualitative comparison we plot the linear regression results (red line), and the Pearson correlation coefficient $p$. \textcolor{black}{Note: we only plot one genuine and one impostor score (randomly chosen) for each of the $1$,$000$ subjects to improve the visualization of the results.}}
\label{Editdistances}
\end{figure*}

In Fig. \ref{Editdistances} we plot the test scores (Euclidean distances) employed in one-shot scenario ($G = 1$ enrollment sequence per subject, $M = 50$ keystrokes per sequence, $k = 1$,$000$ test subjects) versus the Levenshtein distance between the gallery and the query sample that produced the test score (i.e. $d_E(\textbf{f}(\textbf{x}_{g}),\textbf{f}(\textbf{x}_{q}))$ vs. $d_L(\textbf{x}_{g},\textbf{x}_{q})$). To provide a quantitative comparison, we also calculate the Pearson coefficient $p$ and the Linear Regression response as a measure of correlation between both distances (smaller slope indicates a weaker relationship). In mobile scenarios (Fig. \ref{Editdistances} down) we can observe a significant correlation (i.e higher slope in the Linear Regression response and high $p$ value) between the Levenshtein distances and the test scores: genuine distance scores show lower Levenshtein distances (i.e. more similar typed text) than the impostor ones, and therefore, this metric provides us some clues about the possibility that TypeNet models in the mobile scenario could be using the similarity of linguistic expressions or keys pressed between the gallery and the query samples to recognize subjects. These results suggest us that the TypeNet models trained in the mobile scenario may be performing worse than in the desktop scenario, among other factors, because mobile TypeNet embeddings show a significant dependency to the entry text. On the other hand, in desktop scenarios (Fig. \ref{Editdistances} up) this correlation is not present (i.e. the small slope in the Linear Regression response and $p \sim 0$) between test scores and Levenshtein distances, suggesting that the embedding vector produced by TypeNet models trained with the desktop dataset are largely independent of the input text.
%-------------------------------------------------------------------------
\section{Conclusions and Future Work}
\label{conclusions}
We have presented new free-text keystroke biometrics systems based on an RNN architecture trained with \textcolor{black}{different learning strategies and evaluated over 4 public databases. We present a comprehensive performance analysis including authentication and identification results obtained at very large scale. These experiments comprise more than} $136$ million keystrokes from $168$,$000$ subjects captured on desktop keyboards and  $60$,$000$ subjects captured on mobile devices with more than $63$ million keystrokes. Deep neural networks have shown to be effective in face recognition tasks when scaling up to hundreds of thousands of identities \cite{kemelmacher2016megaface}. The same capacity has been shown by TypeNet models in free-text keystroke biometrics. 

In all authentication scenarios evaluated in this work, the models trained with triplet loss have shown a superior performance, especially when there are many subjects but few enrollment samples per subject. The results achieved in this work outperform previous state-of-the-art algorithms. Our results range from  $17.2\%$ to $1.2\%$ EER in desktop and from $17.7\%$ to $6.3\%$ EER in mobile scenarios depending on the amount of subject data enrolled. A good balance between performance and the amount of enrollment data per subject is achieved with $5$ enrollment sequences and $50$ keystrokes per sequence, which yields an EER of $2.2/9.2\%$ (desktop/mobile) for $1$,$000$ test subjects. These results suggest that our approach achieves error rates close to those achieved by the state-of-the-art fixed-text algorithms \cite{2016_IEEEAccess_KBOC_Aythami}, within $\sim$$5\%$ of error rate even when the enrollment data is scarce.

Scaling up the number of test subjects does not significantly affect the performance: the EER in the desktop scenario increases only $5\%$ in relative terms with respect to the previous $2.2\%$ when scaling up from $1$,$000$ to $100$,$000$ test subjects, while in the mobile scenario decays up to $15\%$ the EER in relative terms. Evidence of the EER stabilizing around $10$,$000$ subjects demonstrates the potential of this architecture to perform well at large scale. However, the error rates of both models increase in the cross-device interoperability scenario. Evaluating the TypeNet model trained in the desktop scenario with the mobile dataset the EER increases from $2.2\%$ to $13.7\%$, and from $9.2\%$ to $21.4\%$ for the TypeNet model trained with the mobile dataset when testing with the desktop dataset. A solution based on a mixture model trained with samples from both datasets outperforms the previous TypeNet models in the cross-device scenario but with significantly worse results compared to single-device development and testing. \textcolor{black}{When testing the generalization capacity of the proposed models with the Buffalo and Clarkson II keystroke datasets, TypeNet is able to maintain a competitive performance (between $7.6\%$ and $17.2\%$ of EER for the best scenario) without any kind of transfer learning or retraining, demonstrating the potential of TypeNet models to generalize well in other databases acquired under similar conditions. However, the performance decreased quickly when testing with databases acquired with different conditions or devices (e.g. touchscreen sensors).}  

In addition to authentication results, identification experiments have been also conducted. In this case, TypeNet models trained with triplet loss have shown again a superior performance in all ranks evaluated. For Rank-$1$, TypeNet models trained with triplet loss have an accuracy of $67.4/25.5\%$ (desktop/mobile) with a background size of $\mathfrak{B}=1$,$000$ identities, meanwhile previous related works barely achieve $6.5\%$ accuracy. For Rank-$50$, the TypeNet model trained with triplet loss achieves almost $100\%$ accuracy in the desktop scenario and up to $87.5\%$ in the mobile one. The results are improved when using auxiliary-data to realize a pre-screening of the initial list of gallery profiles (e.g. country, language), showing the potential of TypeNet models to perform great not only in authentication, but also in identification tasks. Finally we have demonstrated that the text-entry dependencies in TypeNet models are irrelevant in desktop scenarios, although in mobile scenarios the TypeNet models have some correlation between the input text typed and the performance achieved.

For future work, we will improve the way training pairs/triplets are chosen in Siamese/Triplet training. Currently, the pairs are chosen randomly; however, recent work has shown that choosing \textit{hard pairs} during the training phase can improve the quality of the embedding feature vectors \cite{Wu}. We will also explore improved learning architectures based on a combination of short- and long-term modeling, which has demonstrated to be very useful for modeling behavioral biometrics \cite{2021_AAAI_DeepWriteSYN_Tolosana}.

In addition, we plan to investigate alternate ways to combine the multiple sources of information \cite{2018_INFFUS_MCSreview2_Fierrez} originated in the proposed framework, e.g., the multiple distances in Equation~(\ref{score}). Integration of keystroke data with other information captured at the same time in desktop \cite{2020_AAAI_edBB_JH} and mobile acquisition \cite{2019_MULEA_Acien_MultiLock} will also be explored.

Finally, the proposed TypeNet models will be valuable beyond user authentication and identification, for applications related to human behavior analysis like profiling \cite{2019_IETB_DetectChildTouch_Acien}, bot detection \cite{2021_EAAI_BeCAPTCHA_Acien}, and e-health \cite{Giancardo2016}.

\ifCLASSOPTIONcompsoc
  % The Biometrics Council usually uses the plural form
  \section*{Acknowledgments}
  
\else
  \section*{Acknowledgment}
\fi
This work has been supported by projects: TRESPASS-ETN (MSCA-ITN-2019-860813), PRIMA (MSCA-ITN-2019-860315), BIBECA (RTI2018-101248-B-I00 MINECO), edBB (UAM), and Instituto de Ingenieria del Conocimiento (IIC). A. Acien is supported by a FPI fellowship from the Spanish MINECO.

\ifCLASSOPTIONcaptionsoff
  \newpage
\fi

\bibliographystyle{IEEEtran}
\bibliography{bibliography.bib}



%\bibliographystyle{IEEEtran}


\begin{IEEEbiography}[{\includegraphics[width=1in,height=1.25in,clip,keepaspectratio]{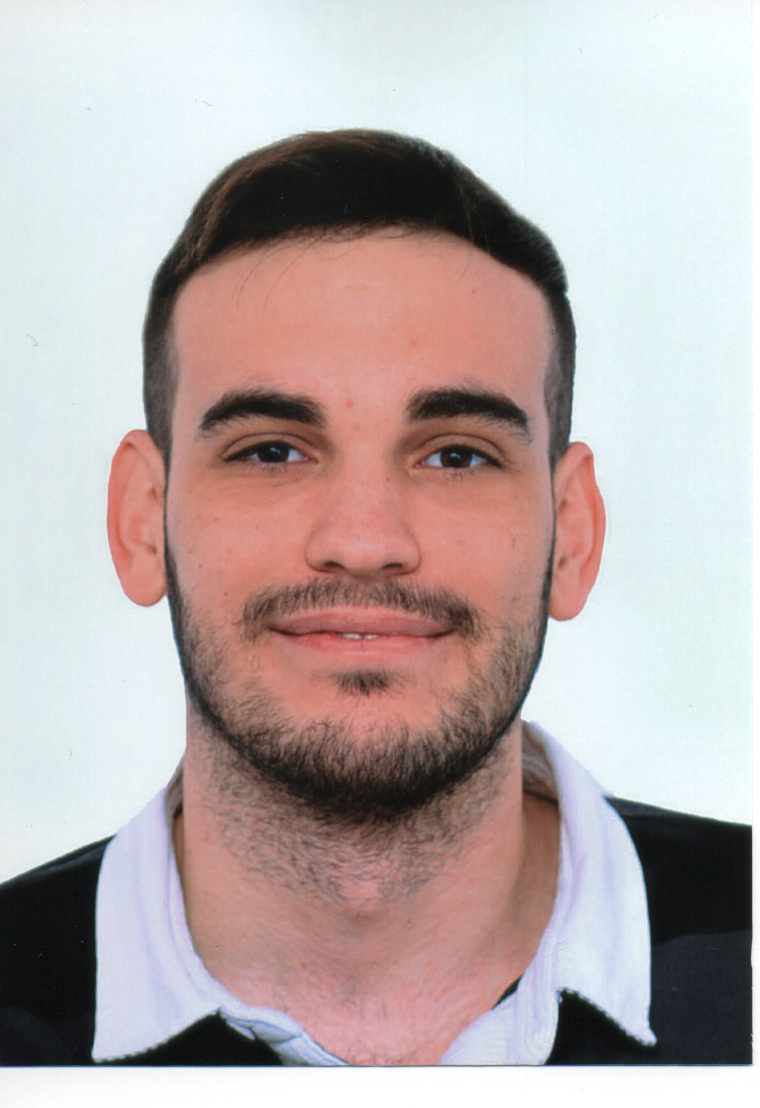}}]{Alejandro Acien}
Alejandro Acien received the MSc in Electrical Engineering in 2015 from Universidad Autonoma de Madrid. In October 2016, he joined the Biometric Recognition Group - ATVS at the Universidad Autonoma de Madrid, where he is currently collaborating as an assistant researcher pursuing the PhD degree. The research activities he is currently working in Behaviour Biometrics, Human-Machine Interaction, Cognitive Biometric Authentication, Machine Learning and Deep Learning.
\end{IEEEbiography}

\begin{IEEEbiography}[{\includegraphics[width=1in,height=1.25in,clip,keepaspectratio]{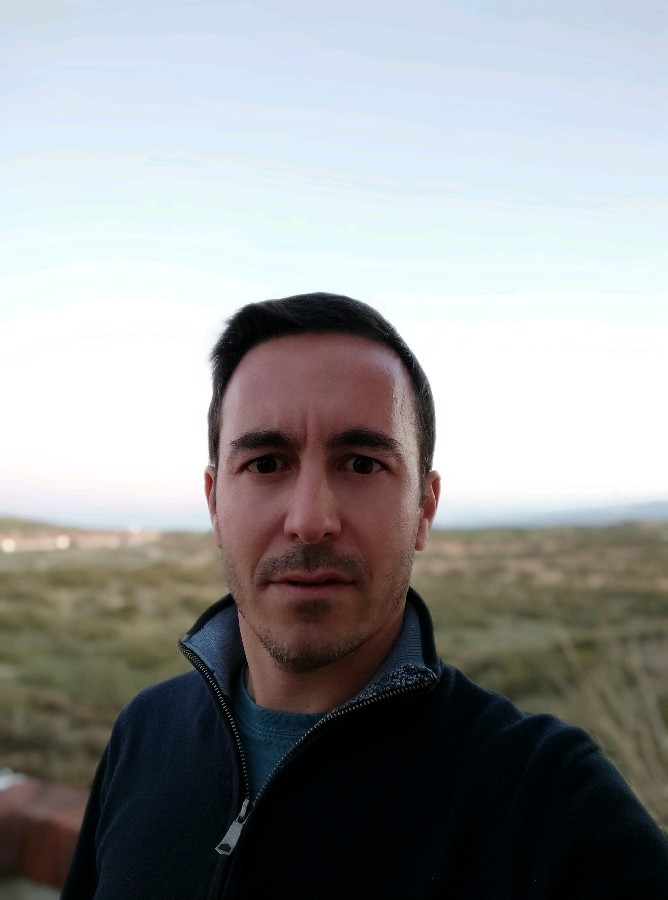}}]{Aythami Morales}
Aythami Morales received his M.Sc. degree in Telecommunication Engineering in 2006 from ULPGC. He received his Ph.D degree from ULPGC in 2011. He performs his research works in the BiDA Lab at Universidad Autónoma de Madrid, where he is currently an Associate Professor. He has performed research stays at the Biometric Research Laboratory at Michigan State University, the Biometric Research Center at Hong Kong Polytechnic University, the Biometric System Laboratory at University of Bologna and Schepens Eye Research Institute. His research interests include pattern recognition, machine learning, trustworthy AI, and biometrics. He is author of more than 100 scientific articles published in international journals and conferences, and 4 patents. He has received awards from ULPGC, La Caja de Canarias, SPEGC, and COIT. He has participated in several National and European projects in collaboration with other universities and private entities such as ULPGC, UPM, EUPMt, Accenture, Unión Fenosa, Soluziona, BBVA.
\end{IEEEbiography}

\begin{IEEEbiography}[{\includegraphics[width=1in,height=1.25in,clip,keepaspectratio]{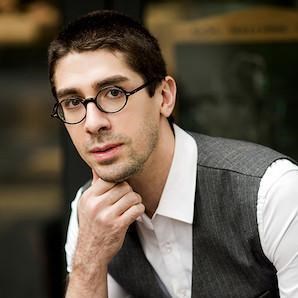}}]{John V. Monaco}
Dr. Monaco is an Assistant Professor in the Computer Science Department at the Naval Postgraduate School in Monterey, CA. His research focuses on user and device fingerprinting, security and privacy in human-computer interaction, and the development of neural-inspired computer architectures. Dr. Monaco is the recipient of Best Paper Awards at the 2020 Conference on Human Factors in Computing Systems and the 2017 International Symposium on Circuits and Systems. His work is supported by the National Reconnaissance Office and the Army Network Enterprise Technology Command.
\end{IEEEbiography}

\begin{IEEEbiography}[{\includegraphics[width=1in,height=1.25in,clip,keepaspectratio]{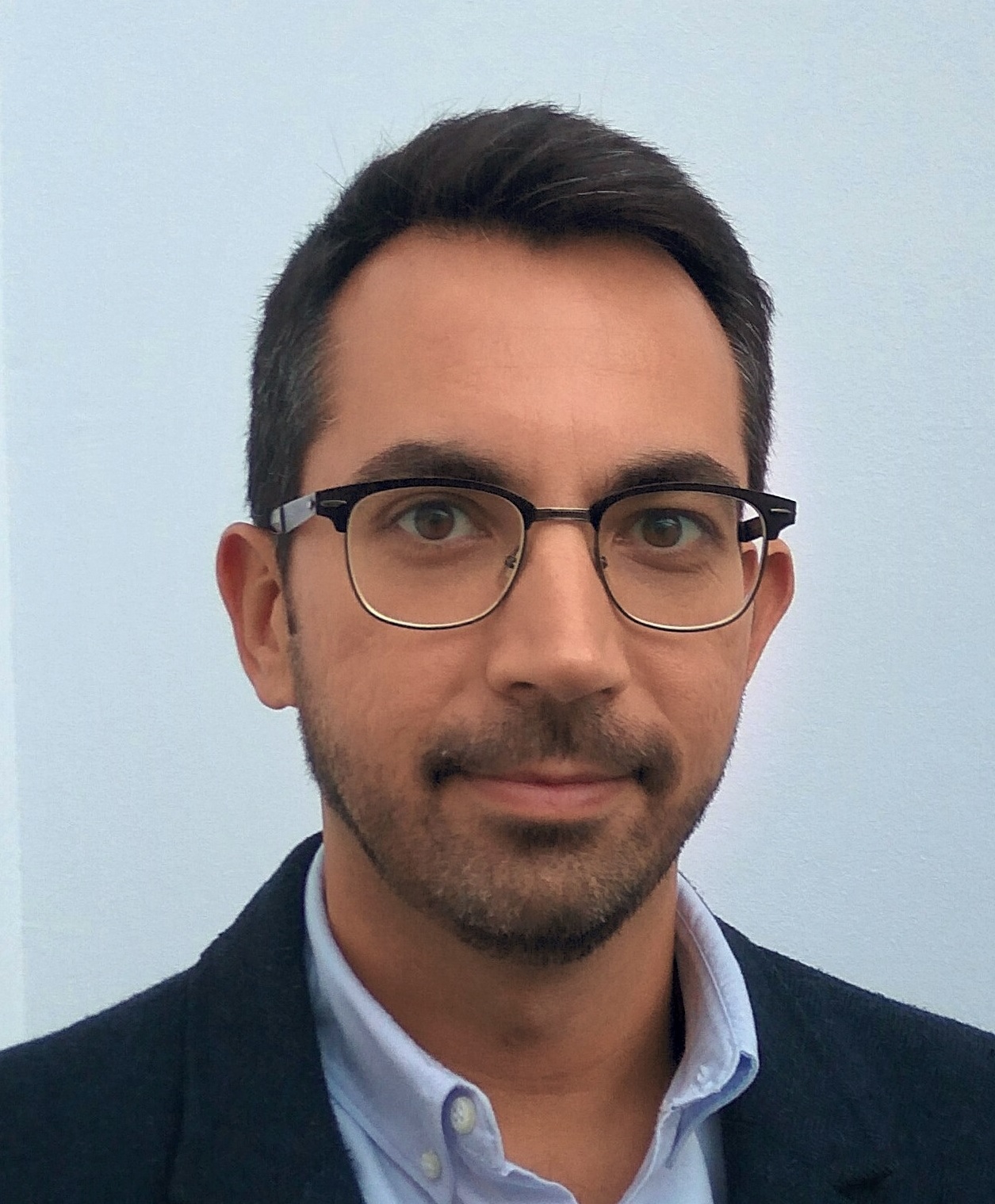}}]{Ruben Vera-Rodriguez}
Ruben Vera-Rodriguez received the M.Sc. degree in telecommunications engineering from Universidad de Sevilla, Spain, in 2006, and the Ph.D. degree in electrical and electronic engineering from Swansea University, U.K., in 2010. Since 2010, he has been affiliated with the Biometric Recognition Group, Universidad Autonoma de Madrid, Spain, where he is currently an Associate Professor since 2018. His research interests include signal and image processing, pattern recognition, HCI, and biometrics, with emphasis on signature, face, gait verification and forensic applications of biometrics. Ruben has published over 100 scientific articles published in international journals and conferences. He is actively involved in several National and European projects focused on biometrics. Ruben has been Program Chair for the IEEE 51st International Carnahan Conference on Security and Technology (ICCST) in 2017; the 23rd Iberoamerican Congress on Pattern Recognition (CIARP 2018) in 2018; and the International Conference on Biometric Engineering and Applications (ICBEA 2019) in 2019.
\end{IEEEbiography}

\begin{IEEEbiography}[{\includegraphics[width=1in,height=1.25in,clip,keepaspectratio]{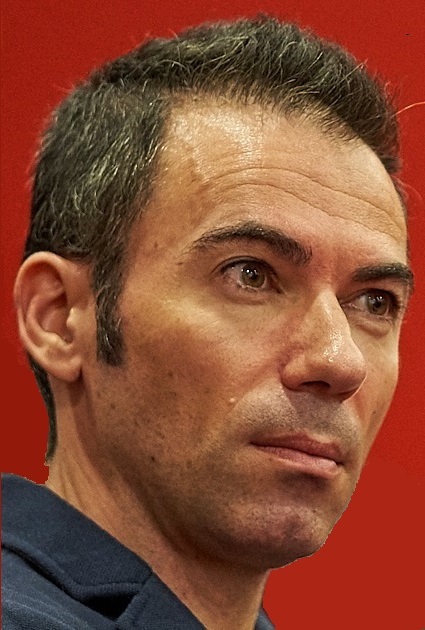}}]{Julian Fierrez}
Julian FIERREZ received the MSc and the PhD degrees from Universidad Politecnica de Madrid, Spain, in 2001 and 2006, respectively. Since 2004 he is at Universidad Autonoma de Madrid, where he is Associate Professor since 2010. His research is on signal and image processing, AI fundamentals and applications, HCI, forensics, and biometrics for security and human behavior analysis. He is Associate Editor for Information Fusion, IEEE Trans. on Information Forensics and Security, and IEEE Trans. on Image Processing. He has received best papers awards at AVBPA, ICB, IJCB, ICPR, ICPRS, and Pattern Recognition Letters; and several research distinctions, including: EBF European Biometric Industry Award 2006, EURASIP Best PhD Award 2012, Miguel Catalan Award to the Best Researcher under 40 in the Community of Madrid in the general area of Science and Technology, and the IAPR Young Biometrics Investigator Award 2017. Since 2020 he is member of the ELLIS Society.
\end{IEEEbiography}


%\vfill


\end{document}